\RequirePackage{fix-cm}
\documentclass[twocolumn]{svjour3}          
\smartqed  
\usepackage{graphicx}
\usepackage{multirow}
\usepackage{hyperref}
\usepackage{amsmath}
\usepackage{cite}
\usepackage{floatrow}

\begin{document}

\title{Pedestrian Alignment Network for \\Large-scale Person Re-identification
}


\author{Zhedong Zheng \and 
	    Liang Zheng \and 
		Yi Yang 
}


\institute{Zhedong Zheng \and Liang Zheng \and Yi Yang \at
			University of Technology Sydney, Australia \\
            \email{zdzheng12@gmail.com} 
}

\date{Received: date / Accepted: date}

\maketitle

\begin{abstract}
Person re-identification (person re-ID) is mostly viewed as an image retrieval problem. This task aims to search a query person in a large image pool. In practice, person re-ID usually adopts automatic detectors to obtain cropped pedestrian images. However, this process suffers from two types of detector errors: excessive background and part missing. Both errors deteriorate the quality of pedestrian alignment and may compromise pedestrian matching due to the position and scale variances.
To address the misalignment problem, we propose that alignment can be learned from an identification procedure. We introduce the pedestrian alignment network (PAN) which allows discriminative embedding learning and pedestrian alignment without extra annotations.  
Our key observation is that when the convolutional neural network (CNN) learns to discriminate between different identities, the learned feature maps usually exhibit strong activations on the human body rather than the background. 
The proposed network thus takes advantage of this attention mechanism to adaptively locate and align pedestrians within a bounding box.
Visual examples show that pedestrians are better aligned with PAN.
Experiments on three large-scale re-ID datasets confirm that PAN improves the discriminative ability of the feature embeddings and yields competitive accuracy with the state-of-the-art methods. \footnote{ The project website of this paper is \url{https://github.com/layumi/Pedestrian_Alignment}.}

\end{abstract}

\newcommand{\etal}{\mbox{\emph{et al.\ }}}
\newcommand{\ie}{\mbox{\emph{i.e.,\ }}}
\section{Introduction}
%
%
%
%

Person re-identification (person re-ID) aims at spotting the target person in different cameras, and is mostly viewed as an image retrieval problem, \emph{i.e.,} searching for the query person in a large image pool (gallery). Recent progress mainly consists in the discriminatively learned embeddings using the convolutional neural network (CNN) on large-scale datasets. The learned embeddings extracted from the fine-tuned CNNs are shown to outperform the hand-crafted features \cite{zheng2016survey,xiao2016learning,zhong2017re}. 



\begin{figure}[t]
\begin{center}
   \includegraphics[width=1\linewidth]{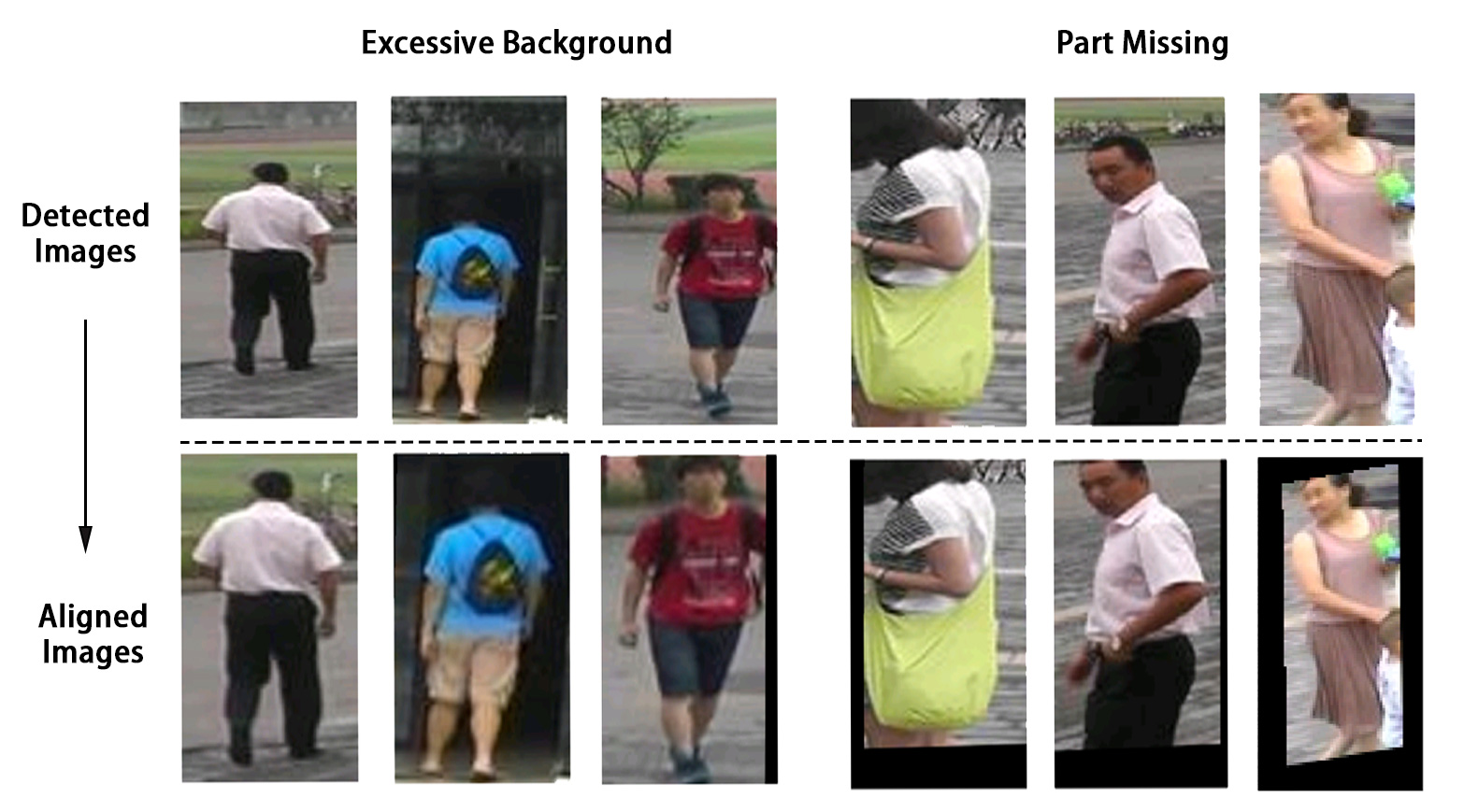}
\end{center}
   \caption{Sample images influenced by detector errors (the first row) which are aligned by the proposed method (the second row). Two types of errors are shown: excessive background and part missing. We show that the pedestrian alignment network (PAN) corrects the misalignment problem by 1) removing extra background or 2) padding zeros to the image borders. PAN reduces the scale and position variance, and the aligned output thus benefit the subsequent matching step.
   }
\label{fig:1}
\end{figure}

Among the many influencing factors, misalignment is a critical one in person re-ID. This problem arises due to the usage of pedestrian detectors. In realistic settings, the hand-drawn bounding boxes, existing in some previous datasets such as VIPER \cite{gray2007evaluating}, CUHK01 \cite{li2012human} and CUHK02 \cite{li2013locally}, are infeasible to acquire when millions of bounding boxes are to be generated. So recent large-scale benchmarks such as CUHK03 \cite{li2014deepreid}, Market1501 \cite{zheng2015scalable} and MARS \cite{zheng2016mars} adopt the Deformable Part Model (DPM) \cite{felzenszwalb2008discriminatively} to automatically detect pedestrians. This pipeline largely saves the amount of labeling effort and is closer to realistic settings. However,  when detectors are used, detection errors are inevitable, which may lead to two common noisy factors: excessive background and part missing. For the former, the background may take up a large proportion of a detected image. For the latter, a detected image may contain only part of the human body, \emph{i.e.,} with missing parts (see Fig. \ref{fig:1}). 

Pedestrian alignment and re-identification are two connected problems. When we have the identity labels of the pedestrian bounding boxes, we might be able to find the optimal affine transformation that contains the most informative visual cues to discriminate between different identities. With the affine transformation, pedestrians can be better aligned. Furthermore, with superior alignment, more discriminative features can be learned, and the pedestrian matching accuracy can, in turn, be improved.

Motivated by the above-mentioned aspects, we propose to incorporate pedestrian alignment into an identification re-ID architecture, yielding the pedestrian alignment network (PAN). Given a pedestrian detected image, this network simultaneously learns to re-localize the person and categorize the person into pre-defined identities. Therefore, PAN takes advantage of the complementary nature of person alignment and re-identification. 

In a nutshell, the training process of PAN is composed of the following components: 1) a network to predict the identity of an input image, 2) an affine transformation to be estimated which re-localizes the input image, and 3) another network to predict the identity of the re-localized image. For components 1) and 3), we use two convolutional branches called the base branch and alignment branch, to respectively predict the identity of the original image and the aligned image. Internally, they share the low-level features and during testing are concatenated at the fully-connected (FC) layer to generate the pedestrian descriptor. In component 2), the affine parameters are estimated using the feature maps from the high-level convolutional layer of the base branch. The affine transformation is later applied on the lower-level feature maps of the base branch. In this step, we deploy a differentiable localization network: spatial transformer network (STN) \cite{jaderberg2015spatial}. With STN, we can 1) crop the detected images which may contain too much background or 2) pad zeros to the borders of images with missing parts. As a result, we reduce the impact of scale and position variances caused by misdetection and thus make pedestrian matching more precise. 

Note that our method addresses the misalignment problem caused by detection errors, while the commonly used patch matching strategy aims to discover matched local structures in well-aligned images. For methods that use patch matching, 
it is assumed that the matched local structures locate in the same horizontal stripe \cite{li2014deepreid,yi2014deep,zhao2013person,zhao2014learning,liao2015person,cheng2016person} or square neighborhood \cite{ahmed2015improved}.
Therefore, these algorithms are robust to some small spatial variance, \emph{e.g.,} position and scale. However, when misdetection happens, due to the limitation of the search scope, this type of methods may fail to discover the matched structures, and the risk of part mismatching may be high. Therefore, regarding the problem to be solved, the proposed method is significantly different from this line of works \cite{li2014deepreid,yi2014deep,ahmed2015improved,zhao2013person,zhao2014learning,liao2015person,cheng2016person}. We speculate that our method is a good complementary step for those using part matching. 

Our contributions are summarized as follows:
\begin{itemize}
\item We propose the pedestrian alignment network (PAN), which simultaneously aligns pedestrians within images and learns pedestrian descriptors. Except for the identity label, we do not need any extra annotation;

\item We observe that the manually cropped images are not as perfect as preassumed to be. We show that our network also improves the re-ID performance on the hand-drawn datasets which are considered to have decent person alignment.

\item We achieve competitive accuracy compared to the state-of-the-art methods on three large-scale person re-ID datasets (Market-1501\cite{zheng2015scalable}, CUHK03 \cite{li2014deepreid} and DukeMTMC-reID \cite{zheng2017unlabeled}).
\end{itemize}

The rest of this paper is organized as follows. Section \ref{relatedwork} reviews and discusses related works. Section \ref{method} illustrates the proposed method in detail. Experimental results and comparisons on three large-scale person re-ID datasets are discussed in Section \ref{experiments}, followed by conclusions in Section \ref{conclusion}.

\section{Related work}  \label{relatedwork}
Our work aims to address two tasks: person re-identification (person re-ID) and person alignment jointly. In this section, we review the relevant works in these two domains. 

\subsection{Hand-crafted Systems for Re-ID} 
Person re-ID needs to find the robust and discriminative features among different cameras. 
Several pioneering approaches have explored person re-ID by extracting local hand-crafted features such as LBP \cite{mignon2012pcca}, Gabor \cite{prosser2010person} and LOMO \cite{liao2015person}. In a series of works by \cite{zhao2014learning,zhao2013person,zhao2013unsupervised}, the 32-dim LAB color histogram and the 128-dim SIFT descriptor are extracted from each $10\times10$ patches. \cite{zheng2015scalable} use color name descriptor for each local patch and aggregate them into a global vector through the Bag-of-Words model. Approximate nearest neighbor search \cite{wang2012query} is employed for fast retrieval but accuracy compromise. \cite{chen2017exemplar} also deploy several different hand-crafted features extracting from overlapped body patches. Differently, \cite{cheng2011custom} localize the parts first and calculate color histograms for part-to-part correspondences. This line of works is beneficial from the local invariance in different viewpoints.

Besides finding robust feature, metric learning is nontrivial for person re-ID. \cite{kostinger2012large} propose ``KISSME'' based on Mahalanobis distance and formulate the pair comparison as a log-likelihood ratio test. Further, \cite{liao2015person} extend the Bayesian face and KISSME to learn a discriminant subspace with a metric. Aside from the methods using Mahalanobis distance, Prosser \etal apply a set of weak RankSVMs to assemble a strong ranker \cite{prosser2010person}. Gray and Tao propose using the AdaBoost algorithm to fuse different features into a single similarity function \cite{gray2008viewpoint}. \cite{loy2010time} propose a cross canonical correlation analysis for the video-based person re-ID.

\begin{figure*}[t]
\begin{center}
   \includegraphics[width=1\linewidth]{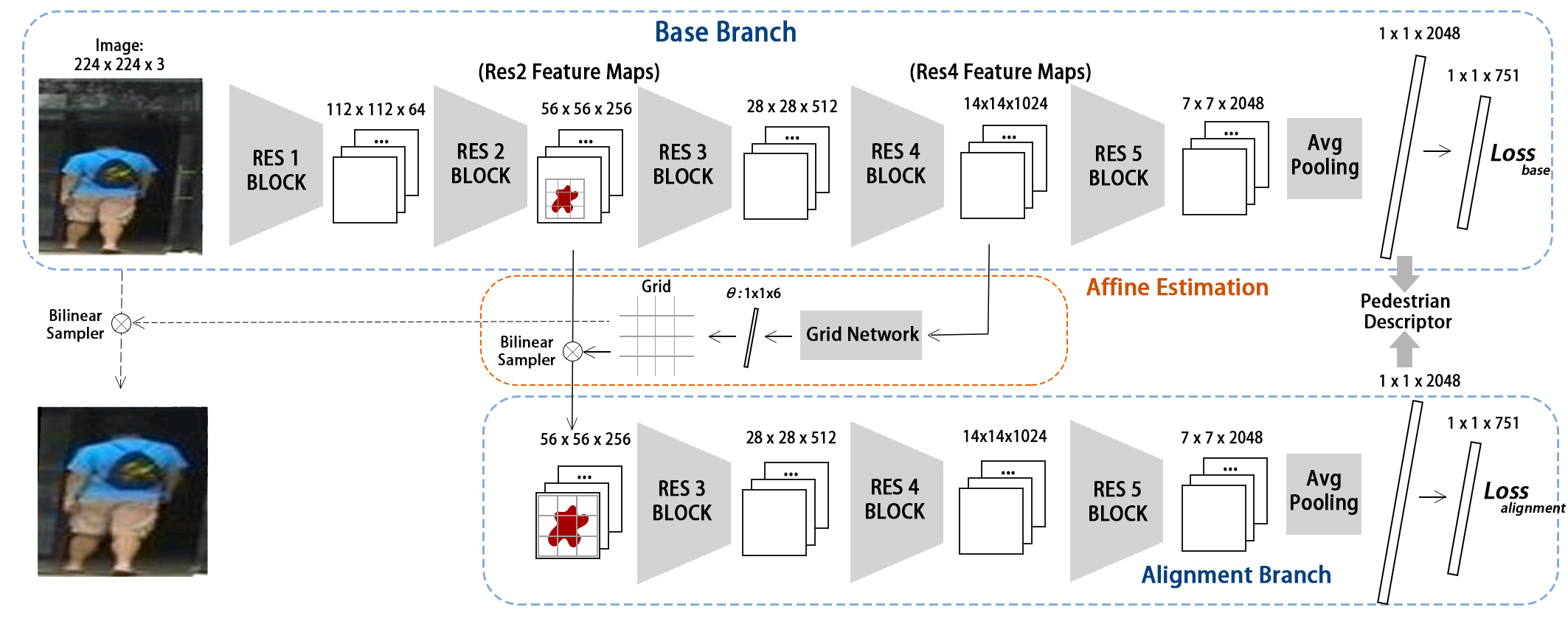}
\end{center}
   \caption{Architecture of the pedestrian alignment network (PAN). It consists of two identification networks (blue) and an affine estimation network (orange). The base branch predicts the identities from the original image. We use the high-level feature maps of the base branch (Res4 Feature Maps) to predict the grid. Then the grid is applied to the low-level feature maps (Res2 Feature Maps) to re-localize the pedestrian (red star). The alignment stream then receives the aligned feature maps to identify the person again. Note that we do not perform alignment on the original images (dotted arrow) as previously done in \cite{jaderberg2015spatial} but directly on the feature maps. In the training phase, the model minimizes two identification losses. In the test phase, we concatenate two $1 \times 1 \times 2048$ FC embeddings to form a $4096$-dim pedestrian descriptor  for  retrieval.}
\label{fig:2}
\end{figure*}
\subsection{Deeply-learned Models for Re-ID}
CNN-based deep learning models have been popular since \cite{krizhevsky2012imagenet} won ILSVRC'12 by a large margin. It extracts features and learns a classifier in an end-to-end system. More recent approaches based on CNN apply spatial constraints by splitting images or adding new patch-matching layers. \cite{yi2014deep} split a pedestrian image into three horizontal parts and respectively trained three part-CNNs to extract features. Similarly, \cite{cheng2016person} split the convolutional map into four parts and fuse the part features with the global feature. \cite{li2014deepreid} add a new layer that multiplies the activation of two images in different horizontal stripes. They use this layer to allow patch matching in CNN explicitly. Later, \cite{ahmed2015improved} improve the performance by proposing a new part-matching layer that compares the activation of two images in neighboring pixels. Besides, \cite{varior2016gated} combine CNN with some gate functions, similar to long-short-term memory (LSTM \cite{hochreiter1997long}) in spirit, which aims to focus on the similar parts of input image pairs adaptively. But it is limited by the computational inefficiency because the input should be in pairs. Similarly, \cite{liu2016end} propose a soft attention-based model to focus on parts and combine CNN with LSTM components selectively; its limitation also consists of the computation inefficiency. 

Moreover, a convolutional network has the high discriminative ability by itself without explicit patch-matching. For person re-ID, \cite{zheng2016person} directly use a conventional fine-tuning approach on Market-1501 \cite{zheng2015scalable} and their performance outperform other recent results. \cite{wu2016enhanced} combine the CNN embedding with hand-crafted features. \cite{xiao2016learning} jointly train a classification model with multiple datasets and propose a new dropout function to deal with the hundreds of identity classes. \cite{wu2016personnet} deepen the network and use filters of smaller size. \cite{lin2017improving} use person attributes as auxiliary tasks to learn more information. \cite{zheng2016discriminatively} propose combining the identification model with the verification model and improve the fine-tuned CNN performance. \cite{ding2015deep} and \cite{hermans2017defense} use triplet samples for training the network which considers the images from the same people and the different people at the same time. Recent work by Zheng \etal combined original training dataset with GAN-generated images and regularized the model \cite{zheng2017unlabeled}. In this paper, we adopt the similar convolutional branches without explicit part-matching layers. It is noted that we focus on a different goal on finding robust pedestrian embedding for person re-identification, and thus our method can be potentially combined with the previous methods to further improve the performance.

\subsection{Objective Alignment}
Face alignment (here refer to the rectification of face misdetection) has been widely studied. \cite{huang2007unsupervised} propose an unsupervised method called funneled image to align faces according to the distribution of other images and improve this method with convolutional RBM descriptor later \cite{huang2012learning}. However, it is not trained in an end-to-end manner, and thus following tasks \ie face recognition take limited benefits from the alignment. On the other hand, several works introduce attention models for task-driven object localization. Jadeburg \etal \cite{jaderberg2015spatial} deploy the spatial transformer network (STN) to fine-grained bird recognition and house number recognition. \cite{johnson2015densecap} combine faster-RCNN \cite{girshick2015fast}, RNN and STN to address the localization and description in image caption. Aside from using STN, Liu \etal use reinforcement learning to detect parts and assemble a strong model for fine-grained recognition \cite{liu2016fully}.

In person re-ID, \cite{baltieri2015mapping} exploits 3D body models to the well-detected images to align the pose but does not handle the misdetection problem. Besides, the work that inspires us the most is ``PoseBox'' proposed by \cite{zheng2017pose}. The PoseBox is a strengthened version of the Pictorial Structures proposed in \cite{cheng2011custom}. PoseBox is similar to our work in that 1) both works aim to solve the misalignment problem, and that 2) the networks have two convolutional streams. Nevertheless, our work differs significantly from PoseBox in two aspects. First, PoseBox employs the convolutional pose machines (CPM) to generate body parts for alignment in advance, while this work learns pedestrian alignment in an end-to-end manner without extra steps. Second, PoseBox can tackle the problem of excessive background but may be less effective when some parts are missing, because CPM fails to detect body joints when the body part is absent. However, our method automatically provides solutions to both problems, \emph{i.e.,} excessive background and part missing. 

\section{Pedestrian Alignment Network} \label{method}
\subsection{Overview of PAN}
Our goal is to design an architecture that jointly aligns the images and identifies the person. The primary challenge is to develop a model that supports end-to-end training and benefits from the two inter-connected tasks. The proposed architecture draws on two convolutional branches and one affine estimation branch to simultaneously address these design constraints. Fig. \ref{fig:2} briefly illustrates our model. 

To illustrate our method, we use the ResNet-50 model \cite{he2016deep} as the base model which is applied on the Market-1501 dataset \cite{zheng2015scalable}. 
Each $Res\_i, i=1,2,3,4,5$ block in Fig. \ref{fig:2} denotes several convolutional layers with batch normalization, ReLU, and optionally max pooling. After each block, the feature maps are down-sampled to be half of the size of the feature maps in the previous block. For example, $Res\_\emph{1}$ down-samples the  width and height of an image from $224 \times 224$ to $112 \times 112$.
In Section \ref{conv} and Section \ref{stn}, we first describe the convolutional branches and affine estimation branches of our model. Then in Section \ref{descriptor} we address the details of a pedestrian descriptor. When testing, we use the descriptor to retrieve the query person. Further, we discuss the re-ranking method as a subsequent processing in Section \ref{re-rank}.

\subsection{Base and Alignment Branches} \label{conv}
\begin{figure}[t]
\begin{center}
   \includegraphics[width=1\linewidth]{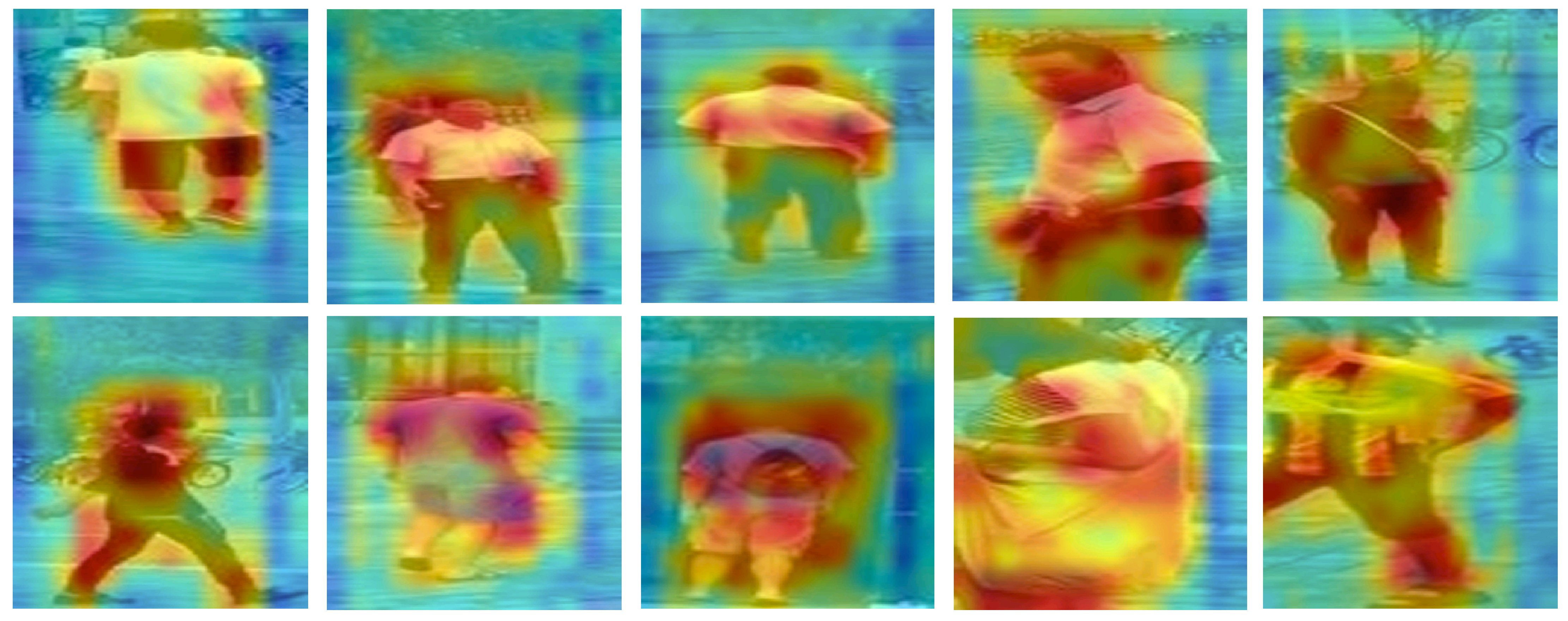}
\end{center}
   \caption{We visualize the Res4 Feature Maps in the base branch. We observe that high responses are mostly concentrated on the pedestrian body. So we use the Res4 Feature Maps to estimate the affine parameters.}
\label{fig:feature_map}
\end{figure}

Recent progress in person re-ID datasets allows the CNN model to learn more discriminative visual representations. There are two main convolutional branches exist in our model, called the base branch and the alignment branch. Both branches are classification networks that predict the identity of the training images. Given an originally detected image, the base branch not only learns to distinguish its identity from the others but also encodes the appearance of the detected image and provides the clues for the spatial localization (see Fig. \ref{fig:feature_map}). The alignment branch shares a similar convolutional network but processes the aligned feature maps produced by the affine estimation branch.

In the base branch, we train the ResNet-50 model \cite{he2016deep}, which consists of five down-sampling blocks and one global average pooling. We deploy the model pre-trained on ImageNet \cite{deng2009imagenet} and remove the final fully-connected (FC) layer. There are $K=751$ identities in the Market-1501 training set, so we add an FC layer to map the CNN embedding of size $1 \times 1 \times 2048$ to $751$ unnormalized probabilities. The alignment branch, on the other hand, is comprised of three ResBlocks and one average pooling layer. We also add an FC layer to predict the multi-class probabilities. The two branches do not share weight. We use $W_1$ and $W_2$ to denote the parameters of the two convolutional branches, respectively.

More formally, given an input image $x$, we use  $p(k|x)$ to denote the probability that the image $x$ belongs to the class $k \in{\{1...K\}}$. Specifically, $p(k|x) = \frac{exp(z_k)}{\sum^{K}_{k=1}exp(z_i)}$. Here $z_i$ is the outputted probability from the CNN model. For the two branches,  the cross-entropy losses are formulated as:
\begin{equation} \label{loss1}
l_{base}(W_1,x,y) = -\sum^{K}_{k=1}(log(p(k|x))q(k|x)),
\end{equation}
\begin{equation} \label{loss2}
l_{align}(W_2,x_{a},y) = -\sum^{K}_{k=1}(log(p(k|x_{a}))q(k|x_{a})),
\end{equation}
where $x_a$ denotes the aligned input. It can be derived from the original input $x_a = T(x)$. Given the label $y$, the ground-truth distribution $q(y|x)=1$ and $q(k|x)=0$ for all $k \neq y$. If we discard the 0 term in Eq. \ref{loss1} and Eq. \ref{loss2}, the losses are equivalent to: 
\begin{equation}
l_{base}(W_1,x,y) = -log(p(y|x)),
\end{equation}
\begin{equation}
l_{align}(W_2,x_{a},y) = -log(p(y|x_{a})).
\end{equation}
Thus, at each iteration, we wish to minimize the total entropy, which equals to maximizing the possibility of the correct prediction. 

\subsection{Affine Estimation Branch} \label{stn} 
To address the problems of excessive background and part missing, the key idea is to predict the position of the pedestrian and do the corresponding spatial transform. When excessive background exists, a cropping strategy should be used; under part missing, we need to pad zeros to the corresponding image borders. Both strategies need to find the parameters for the affine transformation. In this paper, this function is implemented by the affine estimation branch.

The affine estimation branch receives two input tensors of activations $14 \times 14 \times 1024$ and $56 \times 56 \times 256$ from the base branch. We name the two tensors the Res2 Feature Maps and the Res4 Feature Maps, respectively. The Res4 Feature Maps contain shallow feature maps of the original image and reflects the local pattern information. On the other hand, since the Res2 Feature Maps are closer to the  classification layer, it encodes the attention on the pedestrian and semantic cues for aiding identification. The affine estimation branch contains one bilinear sampler and one small network called Grid Network. The Grid Network contains one ResBlock and one average pooling layer. We pass the Res4 Feature Maps through Grid Network to regress a set of 6-dimensional transformer parameters. The learned transforming parameters $\theta$ are used to produce the image grid. The affine transformation process can be written as below,
\begin{equation}
\left( {
\renewcommand{\arraystretch}{1.2}
\begin{array}{c}
x_i^s\\
y_i^s
\end{array}
}\right)
=
\left[ {
\renewcommand{\arraystretch}{1.2}
\begin{array}{ccc}
\theta_{11} & \theta_{12} & \theta_{13}\\
\theta_{21} & \theta_{22} & \theta_{23}
\end{array}
}\right]
\left( {
\renewcommand{\arraystretch}{1.2}
\begin{array}{c}
x^t_i\\
y^t_i\\
1
\end{array}
}\right)
\label{eqn:affine},
\end{equation}
where  $(x^t_i,y^t_i)$ are the target coordinates on the output feature map, and  $(x^s_i,y^s_i)$ are the source coordinates on the input feature maps (Res2 Feature Maps). $\theta_{11}, \theta_{12}, \theta_{21}$ and $\theta_{22}$ deal with the scale and rotation transformation, while
$\theta_{13}$ and $\theta_{23}$ deal with the offset. In this paper, we set the coordinates as: (-1,-1) refer to the pixel on the top left of the image, while (1,1) refer to the bottom right pixel. For example, if $\theta = \left[ {
\renewcommand{\arraystretch}{1.2}
\begin{array}{ccc}
0.8 & 0 & -0.1\\
0 & 0.7 & 0
\end{array}
}\right]$, the pixel value of point $(-1,-1)$ on the output image is equal to that of $(-0.9,-0.7)$ on the original map. 
We use a bilinear sampler to make up the missing pixels, and we assign zeros to the pixels located out of the original range.
So we obtain an injective function from the original feature map $V$ to the aligned output $U$. More formally, we can formulate the equation: 
\begin{multline}
U^c_{(m,n)} = \\
\sum_{x^s}^{H}\sum_{y^s}^{W} V^c_{(x^s,y^s)}max(0,1-|x^t-m|)max(0,1-|y^t-n|).
\end{multline}
$U^c_{(m,n)}$ is the output feature map at location $(m,n)$ in channel $c$, and $V^c_{(x^s,y^s)}$ is the input feature map at location $(x_s,y_s)$ in channel $c$. If $(x_t,y_t)$ is close to $(m,n)$, we add the pixel at $(x_s,y_s)$ according to the bilinear sampling.

In this work, we do not perform pedestrian alignment on the original image; instead, we choose to achieve an equivalent function on the shallow feature maps. By using the feature maps,  we reduce the running time and parameters of the model. This explains why we apply re-localization grid on the feature maps. The bilinear sampler receives the grid, and the feature maps to produce the aligned output $x_a$. We provide some visualization examples in Fig. \ref{fig:feature_map}. Res4 Feature maps are shown. We can observe that through ID supervision, we are able to re-localize the pedestrian and correct misdetections to some extent. 

\subsection{Pedestrian Descriptor} \label{descriptor}
Given the fine-tuned PAN model and an input image $x_i$, the pedestrian descriptor is the weighted fusion of the FC features of the base branch and the alignment branch. That is, we are able to capture the pedestrian characteristic from the original image and the aligned image. In the Section \ref{exp}, the experiment shows that the two features are complementary to each other and improve the person re-ID performance.

In this paper, we adopt a straightforward late fusion strategy, \ie $f_i = g(f_i^1,f_i^2)$. Here $f_i^1$ and $f_i^2$ are the FC descriptors from two types of images, respectively. We reshape the tensor after the final average pooling to a 1-dim vector as the pedestrian descriptor of each branch. The pedestrian descriptor is then written as:
\begin{equation}
f_i = \left[ \alpha|f_i^1|^{\rm T},(1-\alpha)|f_i^2| ^{\rm T}\right]^{\rm T}. 
\end{equation}
The $|\cdot|$ operator denotes an $l^2$-normalization step.
We concatenate the aligned descriptor with the original descriptor, both after $l^2$-normalization.
$\alpha$ is the weight for the two descriptors. If not specified, we simply use $\alpha=0.5$ in our experiments.

\subsection{Re-ranking for re-ID} \label{re-rank}
In this work, we first obtain the rank list $N(q,n) = [x_1,x_2,...x_n]$ by sorting the Euclidean distance of gallery images to the query. Distance is calculated as $D_{i,j} = (f_i-f_j)^2 $, where $f_i$ and $f_j$ are $l_2$-normalized features of image $i$ and $j$, respectively. 
We then perform re-ranking to obtain better retrieval results. Several re-ranking methods can be applied in person re-ID \cite{ye2015ranking,qin2011hello,zhong2017re}. Specifically, we follow a state-of-the-art re-ranking method proposed in \cite{zhong2017re}.

Apart from the Euclidean distance, we additionally consider the Jaccard similarity. To introduce this distance, we first define a robust retrieval set for each image. The k-reciprocal nearest neighbors $R(p,k)$ are comprised of such element that appears in the top-k retrieval rank of the query $p$ while the query is in the top-k rank of the element as well. 
\begin{equation}
R(p,k) = \left\{ x| x\in N(p,k), p \in N(x,k) \right\}
\end{equation}
According to \cite{zhong2017re}, we extend the set $R$ to $R^*$ to include more positive samples. Taking the advantage of set $R^*$, we use the Jaccard similarity for re-ranking. When we use the correctly matched images to conduct the retrieval, we should retrieve a similar rank list as we use the original query. The Jaccard distance can be simply formulated as:
\begin{equation}
D_{similarity} = 1 - \frac{|R^*(q,k)\cap R^*(x_i,k)|}{|R^*(q,k) \cup R^*(x_i,k)|},
\end{equation}
where $|\cdot|$ denotes the cardinality of the set. If $R^*(q,k)$ and $R^*(x_i,k)$ share more elements, $x_i$ is more likely to be a true match. This usually helps us to distinguish some hard negative samples from the correct matches. During testing, this similarity distance is added to the Euclidean distance to re-rank the result. In the experiment, we show that re-ranking further improves our results.

\section{Experiments} \label{experiments}
\begin{figure}[t]
\begin{center}
   \includegraphics[width=1\linewidth]{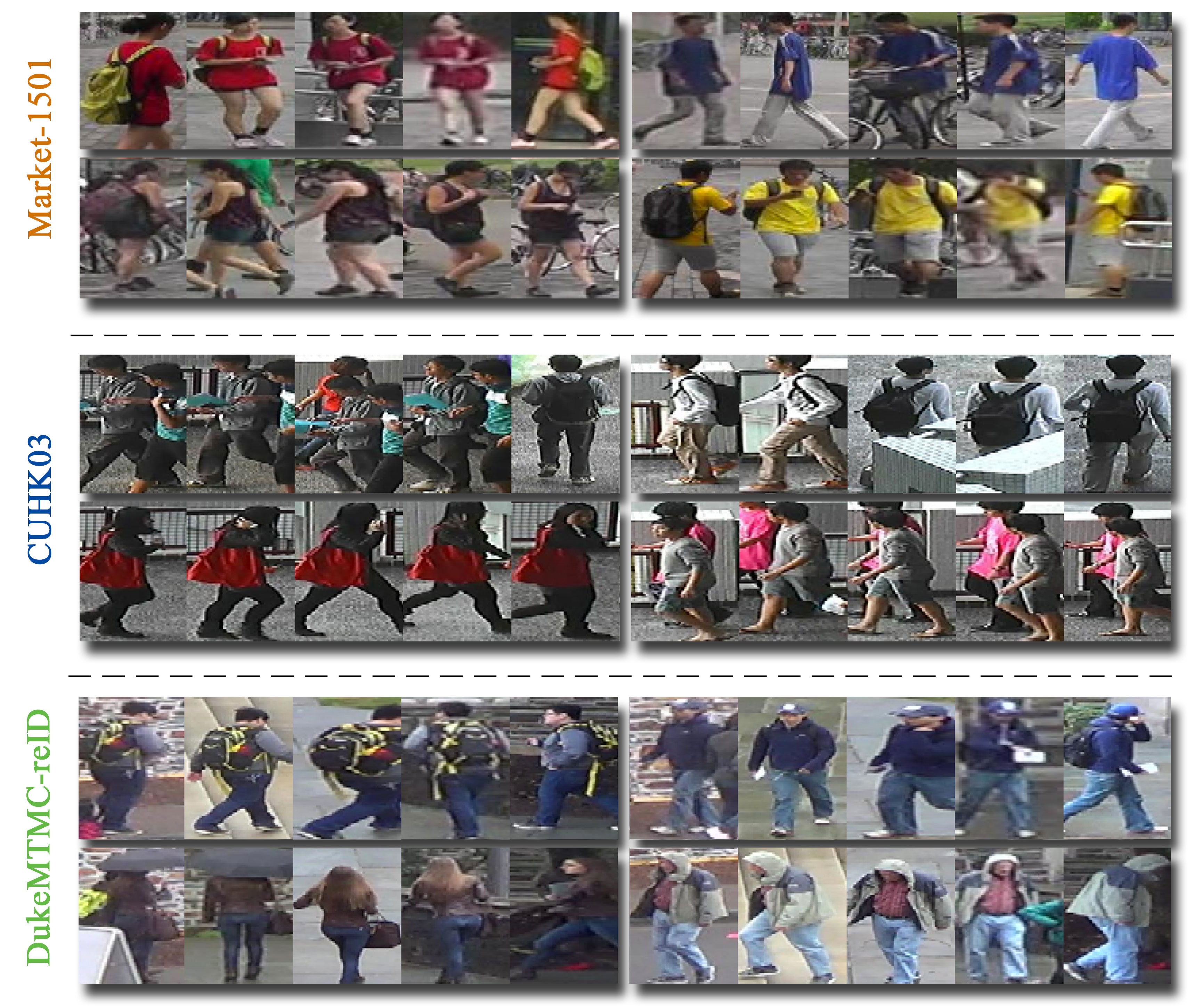}
\end{center}
   \caption{Sample images from Market-1501 \cite{zheng2015scalable}, CUHK03 \cite{li2014deepreid} and DukeMTMC-reID \cite{zheng2017unlabeled}. The three datasets are collected in different scenes with different detection bias.}
\label{fig:dataset}
\end{figure}

In this section, we report the results on three large-scale datasets: Market-1501 \cite{zheng2015scalable}, CUHK03 \cite{li2014deepreid} and DukeMTMC-reID \cite{zheng2017unlabeled}. As for the detector, Market-1501 and CUHK03 (detected) datasets are automatically detected by DPM and face the misdetection problem. It is unknown if the manually annotated images after slight alignment would bring any extra benefit. So we also evaluate the proposed method on the manually annotated images of CUHK03 (labeled) and DukeMTMC-reID, which consist of hand-drawn bounding boxes. As shown in Fig. \ref{fig:dataset}, the three datasets have different characteristics, \ie scene variances, and detection bias.

\begin{table*}[t]
\setlength{\tabcolsep}{2.8pt}
\begin{center}
\begin{tabular}{l|l|cccc|cccc|cccc|cccc}\hline
\multirow{2}{*}{Methods}& \multirow{2}{*}{dim} & \multicolumn{4}{c|}{\emph{Market-1501}}   & \multicolumn{4}{c|}{\emph{CUHK03 (detected)}} &\multicolumn{4}{c|}{\emph{CUHK03 (labeled)}} &\multicolumn{4}{c}{\emph{DukeMTMC-reID}} \\\cline{3-18}
& & 1 & 5 & 20 &mAP & 1 & 5 & 20 & mAP & 1 & 5 & 20& mAP & 1 & 5 & 20 & mAP\\
\hline
 Base &2,048 & 80.17 & 91.69 & 96.59 & 59.14 
 & 30.50 & 51.07 & 71.64 & 29.04 
 & 31.14 & 52.00 & 74.21 & 29.80 
 & 65.22 & 79.13 & 87.75 & 44.99 \\
 Alignment & 2,048 &79.01 & 90.86 & 96.14 & 58.27
 & 34.14 & 54.50 & 72.71 & 31.71
 & 35.29 & 53.64 & 72.43 & 32.90 
 & 68.36 & 81.37 & 88.64 & 47.14\\
\hline
 PAN & 4,096 & 82.81 & 93.53 & 97.06 & 63.35 
 & 36.29 & 55.50 & 75.07 & 34.00 
 & 36.86 & 56.86 & 75.14 & 35.03
 & 71.59 & 83.89 & 90.62 & 51.51\\
\hline
\end{tabular}
\end{center}
\caption{Comparison of different methods on Market-1501, CUHK03 (detected), CUHK03 (labeled) and DukeMTMC-reID. Rank-1, 5, 20 accuracy (\%) and mAP (\%) are shown. Note that the base branch is the same as the classification baseline \cite{zheng2016survey}. We observe consistent improvement of our method over the individual branches on the three datasets.}
\label{table:multi-dataset}
\end{table*}

\subsection{Datasets}
\textbf{Market1501} is a large-scale person re-ID dataset collected in a university campus. It contains 19,732 gallery images, 3,368 query images and 12,936 training images collected from six cameras. There are 751 identities in the training set and 750 identities in the testing set without overlapping. Every identity in the training set has 17.2 photos on average. All images are automatically detected by the DPM detector \cite{felzenszwalb2008discriminatively}. The misalignment problem is common, and the dataset is close to the realistic settings. We use all the 12,936 detected images to train the network and follow the evaluation protocol in the original dataset. There are two evaluation settings. A single query is to use one image of one person as query, and multiple query means to use several images of one person under one camera as a query.

\textbf{CUHK03} contains 14,097 images of 1,467 identities \cite{li2014deepreid}. Each identity contains 9.6 images on average. We follow the new training/testing protocol proposed in \cite{zhong2017re} to evaluate the multi-shot re-ID performance. This setting uses a larger testing gallery and is different from the papers published earlier than \cite{zhong2017re}, such as \cite{liu2016end} and \cite{varior2016siamese}. There are 767 identities in the training set and 700 identities in the testing set (The former setting uses 1,367 IDs for training and the other 100 IDs for testing). Since we usually face a large-scale searching image pool cropped from surveillance videos, a larger testing pool is closer to the realistic image retrieval setting. In the ``detected'' set, all the bounding boxes are produced by DPM; in the ``labeled'' set, the images are all hand-drawn. In this paper, we evaluate our method on ``detected'' and ``labeled'' sets, respectively. If not specified, ``CUHK03'' denotes the detected set, which is more challenging.

\textbf{DukeMTMC-reID} is a subset of the DukeMTMC \cite{ristani2016MTMC} and contains 36,411 images of 1,812 identities shot by eight high-resolution cameras. It is one of the largest pedestrian image datasets. The pedestrian images are cropped from hand-drawn bounding boxes. The dataset consists 16,522 training images of 702 identities, 2,228 query images of the other 702 identities and 17,661 gallery images. It is challenging because many pedestrians are in the similar clothes, and may be occluded by cars or trees. We follow the evaluation protocol in \cite{zheng2017unlabeled}.

\textbf{Evaluation Metrics.} We evaluate our method with rank-1, 5, 20 accuracy and mean average precision (mAP). Here, rank-$i$ accuracy denotes the probability whether one or more correctly matched images appear in top-$i$. If no correctly matched images appear in the top-$i$ of the retrieval list, rank-$i=0$, otherwise rank-$i=1$. We report the mean rank-$i$ accuracy for query images. On the other hand, for each query, we calculate the area under the Precision-Recall curve, which is known as the average precision (AP). The mean of the average precision (mAP) then is calculated, which reflects the precision and recall rate of the performance.

\subsection{Implementation Details} \label{details}
\textbf{ConvNet.} In this work, we first fine-tune the base branch on the person re-ID datasets. Then, the base branch is fixed while we fine-tune the whole network. Specifically, when fine-tuning the base branch, the learning rate decrease from $10^{-3}$ to $10^{-4}$ after 30 epochs. We stop training at the 40th epoch. Similarly, when we train the whole model, the learning rate decrease from $10^{-3}$ to $10^{-4}$ after 30 epochs. We stop training at the 40th epoch. We use mini-batch stochastic gradient descent with a Nesterov momentum fixed to $0.9$ to update the weights. Our implementation is based on the Matconvnet \cite{vedaldi15matconvnet} package. The input images are uniformly resized to the size of $224 \times 224$. In addition, we perform simple data augmentation such as cropping and horizontal flipping following \cite{zheng2016discriminatively}. 

\textbf{STN.} For the affine estimation branch, the network may fall into a local minimum in early iterations. To stabilize training, we find that a small learning rate is useful. We, therefore, use a learning rate of $1 \times 10^{-5}$ for the final convolutional layer in the affine estimation branch. In addition, we set the all $\theta=0$ except that $\theta_{11},\theta_{22}=0.8$ and thus, the alignment branch starts training from looking at the center part of the Res2 Feature Maps. 

\subsection{Evaluation} \label{exp}
\textbf{Evaluation of the ResNet baseline.} We implement the baseline according to the routine proposed in \cite{zheng2016survey}, with the details specified in Section \ref{details}. We report our baseline results in  Table \ref{table:multi-dataset}. The rank-1 accuracy is 80.17\%, 30.50\%, 31.14\% and 65.22\% on Market1501, CUHK03 (detected), CUHK03 (labeled) and DukeMTMC-reID respectively. The baseline model is on par with the network in  \cite{zheng2016discriminatively,zheng2016survey}. In our recent implementation, we use a smaller batch size of 16 and a dropout rate of 0.75. We have therefore obtained a higher baseline rank-1 accuracy 80.17\% on Market-1501 than 73.69\% in the \cite{zheng2016discriminatively,zheng2016survey}. For a fair comparison, we will present the results of our methods built on this new baseline. Note that this baseline result itself is higher than many previous works \cite{barbosa2017looking,varior2016gated,zheng2017pose,zheng2016discriminatively}.

\begin{figure}[t]
\begin{center}
   \includegraphics[width=1\linewidth]{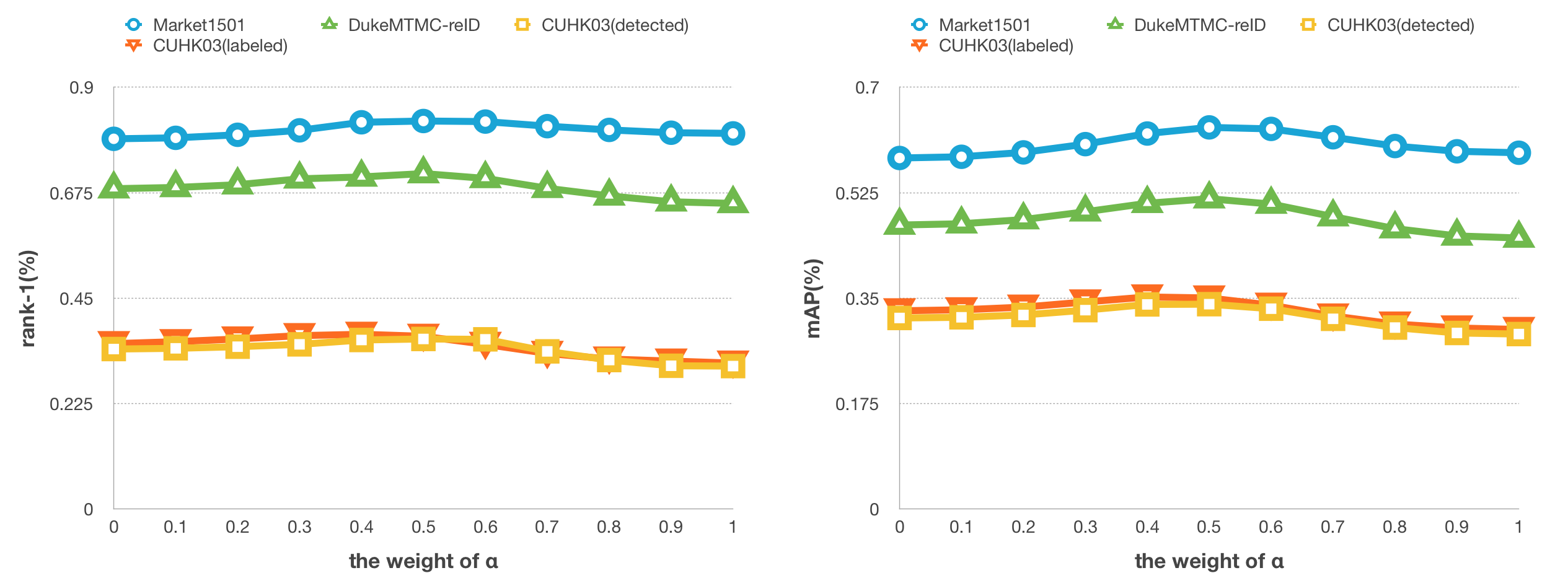}
\end{center}
   \caption{Sensitivity of person re-ID accuracy to parameter $\alpha$. Rank-1 accuracy(\%) and mAP(\%) on three datasets are shown.}
\label{fig:alpha}
\end{figure}

\setlength{\tabcolsep}{15pt}
\begin{table*}
\begin{center}
\begin{tabular}{l|cc|cc}
\hline
\multirow{2}{*}{Method} & \multicolumn{2}{c|}{Single Query} & \multicolumn{2}{c}{Multi. Query}\\
& rank-1 & mAP & rank-1  & mAP \\
\hline
DADM \cite{su2016deep} & 39.4 & 19.6 & 49.0 & 25.8  \\ 
BoW+kissme \cite{zheng2015scalable} & 44.42 & 20.76  & - & -\\
MR-CNN \cite{ustinova2015multiregion}* & 45.58 & 26.11 & 56.59 & 32.26 \\
MST-CNN \cite{liu2016multi} & 45.1 & - & 55.4 & -\\
FisherNet \cite{wu2016deep}* & 48.15 & 29.94 & - & -\\
CAN \cite{liu2016end}* & 48.24 & 24.43  & - & -\\
SL \cite{chen2016similarity} & 51.90 & 26.35  & - & -\\
S-LSTM \cite{varior2016siamese} & - & - & 61.6 & 35.3 \\
DNS \cite{zhang2016learning} & 55.43 & 29.87 & 71.56 & 46.03 \\
Gate Reid \cite{varior2016gated} & 65.88 & 39.55& 76.04 & 48.45 \\
SOMAnet \cite{barbosa2017looking}* & 73.87 & 47.89 & 81.29 & 56.98 \\
PIE \cite{zheng2017pose}* & 78.65 & 53.87 & - & -\\
Verif.-Identif. \cite{zheng2016discriminatively}* & 79.51 & 59.87  & 85.84 & 70.33 \\
SVDNet \cite{sun2017svdnet}* & 82.3 & 62.1 & - & -\\
DeepTransfer \cite{geng2016deep}* & 83.7 & 65.5 & 89.6 & 73.8 \\ 
GAN \cite{zheng2017unlabeled}* & 83.97 & 66.07 & 88.42 & 76.10  \\
APR \cite{lin2017improving}* & 84.29 & 64.67 & - & - \\
Triplet \cite{hermans2017defense}* & 84.92 & 69.14 & 90.53 & 76.42 \\
Triplet+re-rank \cite{hermans2017defense}* & 86.67 & 81.07 & \textbf{91.75} & 87.18\\
\hline
Basel.  & 80.17 & 59.14 & 87.41 & 72.05 \\
Ours & 82.81 & 63.35 & 88.18  & 71.72 \\
Ours+re-rank & 85.78 & 76.56 & 89.79 & 83.79 \\
Ours (GAN) & 86.67 & 69.33 & 90.88 & 76.32\\
Ours (GAN)+re-rank & \textbf{88.57} & \textbf{81.53} & 91.45 & \textbf{87.44}\\
\hline
\end{tabular}
\end{center}
\caption{Rank-1 precision (\%) and mAP (\%) on  Market-1501. We also provide results of the fine-tuned ResNet50 baseline which has the same accuracy with the base branch. * the respective paper is on ArXiv but not published.}
\label{table:mr}
\end{table*}

\textbf{Base branch. vs. alignment branch} To investigate how alignment helps to learn discriminative pedestrian representations, we evaluate the Pedestrian descriptors of the base branch and the alignment branch, respectively. Two conclusions can be inferred. 

First, as shown in Table \ref{table:multi-dataset}, the alignment branch yields higher performance \ie +3.64\% / +4.15\% on the two dataset settings (CUHK03 detected/labeled) and +3.14\% on DukeMTMC-reID,  and achieves a very similar result with the base branch on Market-1501. We speculate that Market-1501 contains more intensive detection errors than the other three datasets and thus, the effect of alignment is limited. 

Second, although the CUHK (labeled) dataset and the DukeMTMC-reID dataset are manually annotated, the alignment branch still improves the performance of the base branch. This observation demonstrates that the manual annotations may not be good enough for the machine to learn a good descriptor. In this scenario, alignment is non-trivial and makes the pedestrian representation more discriminative. 

\textbf{The complementary of the two branches.} As mentioned, the two descriptors capture the different pedestrian characteristic from the original image and the aligned image. We follow the setting in Section \ref{descriptor} and simply combine the two features to form a stronger pedestrian descriptor. The results are summarized in Table \ref{table:multi-dataset}. We observe a constant improvement on the three datasets when we concatenate the two branch descriptors. The fused descriptor improves +2.64\%, +2.15\%, +1.63\% and 3.23\% on Market-1501, CUHK03(detected), CUHK03(labeled) and DukeMTMC-reID, respectively. The two branches are complementary and thus, contain more meaningful information than a separate branch. Aside from the improvement of the accuracy, this simple fusion is efficient sine it does not introduce additional computation.

\textbf{Parameter sensitivity.}
We evaluate the sensitivity of the person re-ID accuracy to the parameter $\alpha$. As shown in Fig. \ref{fig:alpha}, we report the rank-1 accuracy and mAP when tuning the $\alpha$ from 0 to 1. We observe that the change of rank-1 accuracy and mAP are relatively small corresponding to the $\alpha$. Our reported result simply use $\alpha = 0.5$. $\alpha=0.5$ may not be the best choice for a particular dataset. But if we do not foreknow the distribution of the dataset, it is a simple and straightforward choice.

\setlength{\tabcolsep}{5pt}
\begin{table}
\begin{center}
\begin{tabular}{l|cc}
\hline
Method & rank-1 & mAP \\
\hline
BoW+kissme \cite{zheng2015scalable} & 25.13 & 12.17 \\
LOMO+XQDA \cite{liao2015person} & 30.75 & 17.04\\
Gan \cite{zheng2017unlabeled} & 67.68  & 47.13\\
OIM \cite{xiao2017joint} & 68.1 & - \\
APR \cite{lin2017improving} & 70.69 & 51.88 \\
SVDNet \cite{sun2017svdnet} & \textbf{76.7} & 56.8 \\
\hline
Basel. \cite{zheng2017unlabeled} & 65.22 & 44.99\\ 
Ours & 71.59 & 51.51 \\
Ours + re-rank  & 75.94 & \textbf{66.74} \\
\hline
\end{tabular}
\end{center}
\caption{Rank-1 accuracy (\%) and mAP (\%) on DukeMTMC-reID. We follow the evaluation protocol in \cite{zheng2017unlabeled}. We also provide the result of the fine-tuned ResNet50 baseline for fair comparison.}
\label{table:duke}
\end{table}

\setlength{\tabcolsep}{15pt}
\begin{table*}
\begin{center}
\begin{tabular}{l|cc|cc}
\hline
\multirow{2}{*}{Method} & \multicolumn{2}{c|}{Detected} & \multicolumn{2}{c}{Labeled}\\
& rank-1 & mAP & rank-1  & mAP \\
\hline
BoW+XQDA \cite{zheng2015scalable} & 6.36 & 6.39 & 7.93 &7.29 \\
LOMO+XQDA \cite{liao2015person} & 12.8 & 11.5 & 14.8 & 13.6\\
ResNet50+XQDA \cite{zhong2017re} & 31.1 & 28.2 & 32.0 & 29.6\\
ResNet50+XQDA+re-rank \cite{zhong2017re} & 34.7 & 37.4 & 38.1 & 40.3\\
\hline
Basel. & 30.5 & 29.0 & 31.1& 29.8\\
Ours & 36.3 & 34.0 & 36.9 & 35.0 \\
Ours+re-rank & \textbf{41.9} & \textbf{43.8} & \textbf{43.9} & \textbf{45.8} \\
\hline
\end{tabular}
\end{center}
\caption{Rank-1 accuracy (\%) and mAP (\%) on CUHK03 using the new evaluation protocol in \cite{zhong2017re}. This setting uses a larger testing gallery and is different from the papers published earlier than \cite{zhong2017re}, such as \cite{liu2016end} and \cite{varior2016siamese}. There are 767 identities in the training set and 700 identities in the testing set (The former setting uses 1,367 IDs for training and the other 100 IDs for testing). Since we usually face a large-scale searching image pool cropped from surveillance videos, a larger testing pool is more challenging and closer to the realistic image retrieval setting. So we evaluate the proposed method on the ``detected'' and ``labeled'' subsets according to this new multi-shot protocol. We also provide the result of our fine-tuned ResNet50 baseline for fair comparison.}
\label{table:cuhk}
\end{table*}

\begin{figure*}
\floatbox[{\capbeside\thisfloatsetup{capbesideposition={left,top},capbesidewidth=5cm}}]{figure}[\FBwidth]
{\caption{Sample retrieval results on the three datasets. The images in the first column are queries. The retrieved images are sorted according to the similarity score from left to right. For each query, the first row shows the result of baseline \cite{zheng2016survey}, and the second row denotes the results of PAN. The correct and false matches are in the blue and red rectangles, respectively. Images in the rank lists obtained by PAN demonstrate amelioration in alignment. Best viewed when zoomed in.}\label{fig:duke}}
{\includegraphics[width=1.1\linewidth]{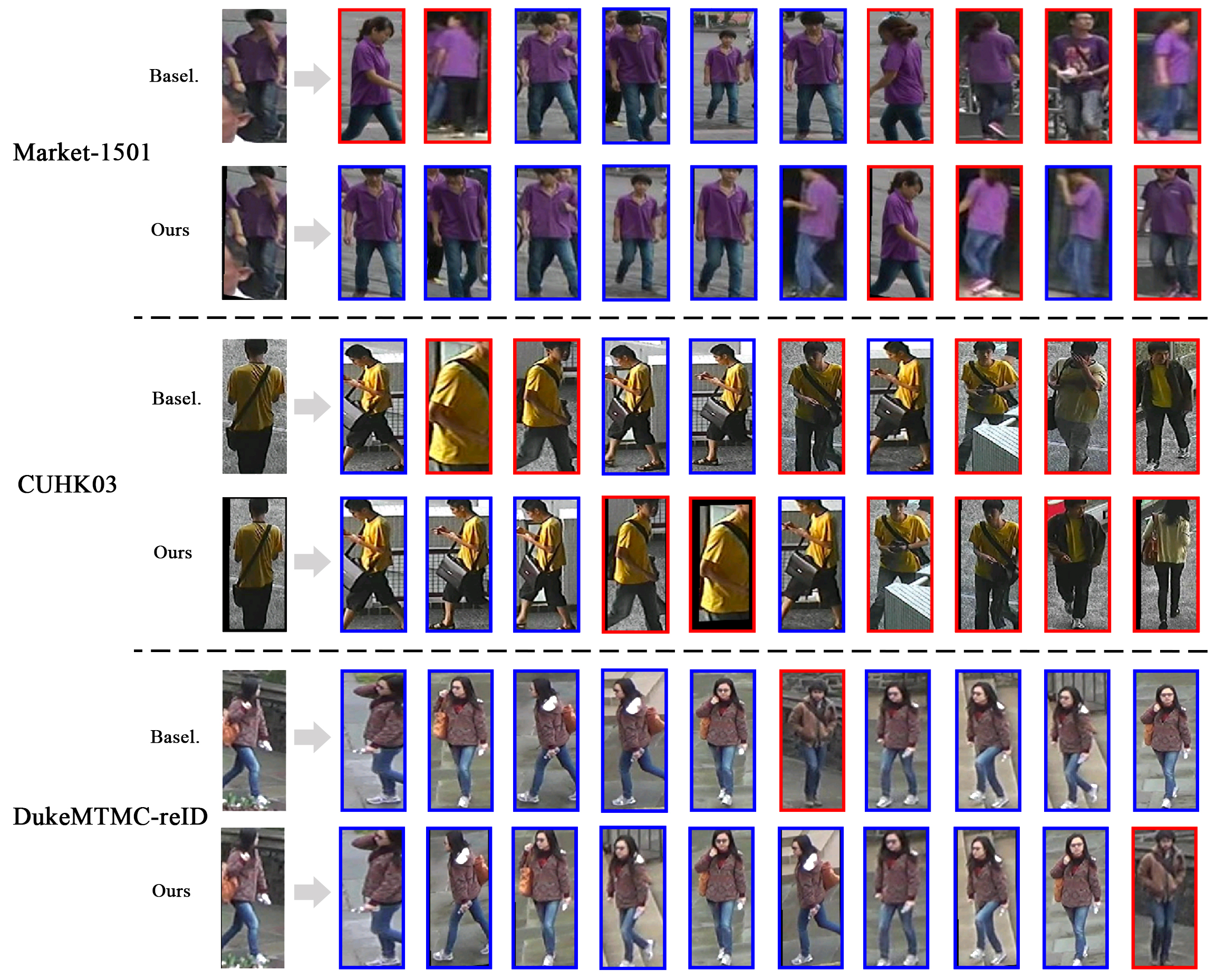}}
\end{figure*}

\textbf{Comparison with the state-of-the-art methods.} We compare our method with the state-of-the-art methods on Market-1501, CUHK03 and DukeMTMC-reID in Table \ref{table:mr}, Table \ref{table:cuhk} and Table \ref{table:duke}, respectively. On  Market-1501, we achieve \textbf{rank-1 accuracy = 85.78\%, mAP = 76.56\%} after re-ranking, which is the best result compared to the published paper, and the second best among all the available results including the arXiv paper. Our model is also adaptive to previous models. One of the previous best results is based on the model regularized by GAN \cite{zheng2017unlabeled}. We combine the model trained on GAN generated images and thus, achieve the state-of-the-art result \textbf{rank-1 accuracy = 88.57\%, mAP = 81.53\%} on Market-1501. 
On CUHK03, we arrive at a competitive result \textbf{rank-1 accuracy = 36.3\%, mAP=34.0\%} on the detected dataset and \textbf{rank-1 accuracy = 36.9\%, mAP = 35.0\%} on the labeled dataset. After re-ranking, we further achieve a state-of-the art result \textbf{rank-1 accuracy = 41.9\%, mAP=43.8\%} on the detected dataset and \textbf{rank-1 accuracy = 43.9\%, mAP = 45.8\%} on the labeled dataset.
On DukeMTMC-reID, we also observe a state-of-the-art result \textbf{rank-1 accuracy = 75.94\% and mAP = 66.74\%} after re-ranking. Despite the visual disparities among the three datasets, \ie scene variance, and detection bias, we show that our method consistently improves the re-ID performance.

As shown in Fig. \ref{fig:duke}, we visualize some retrieval results on the three datasets. Images in the rank lists obtained by PAN demonstrate amelioration in alignment. Comparing to the baseline, true matches which are misaligned originally receive higher ranks, while false matches have lower ranks

\begin{figure*}[t]
\begin{center}
   \includegraphics[width=1\linewidth]{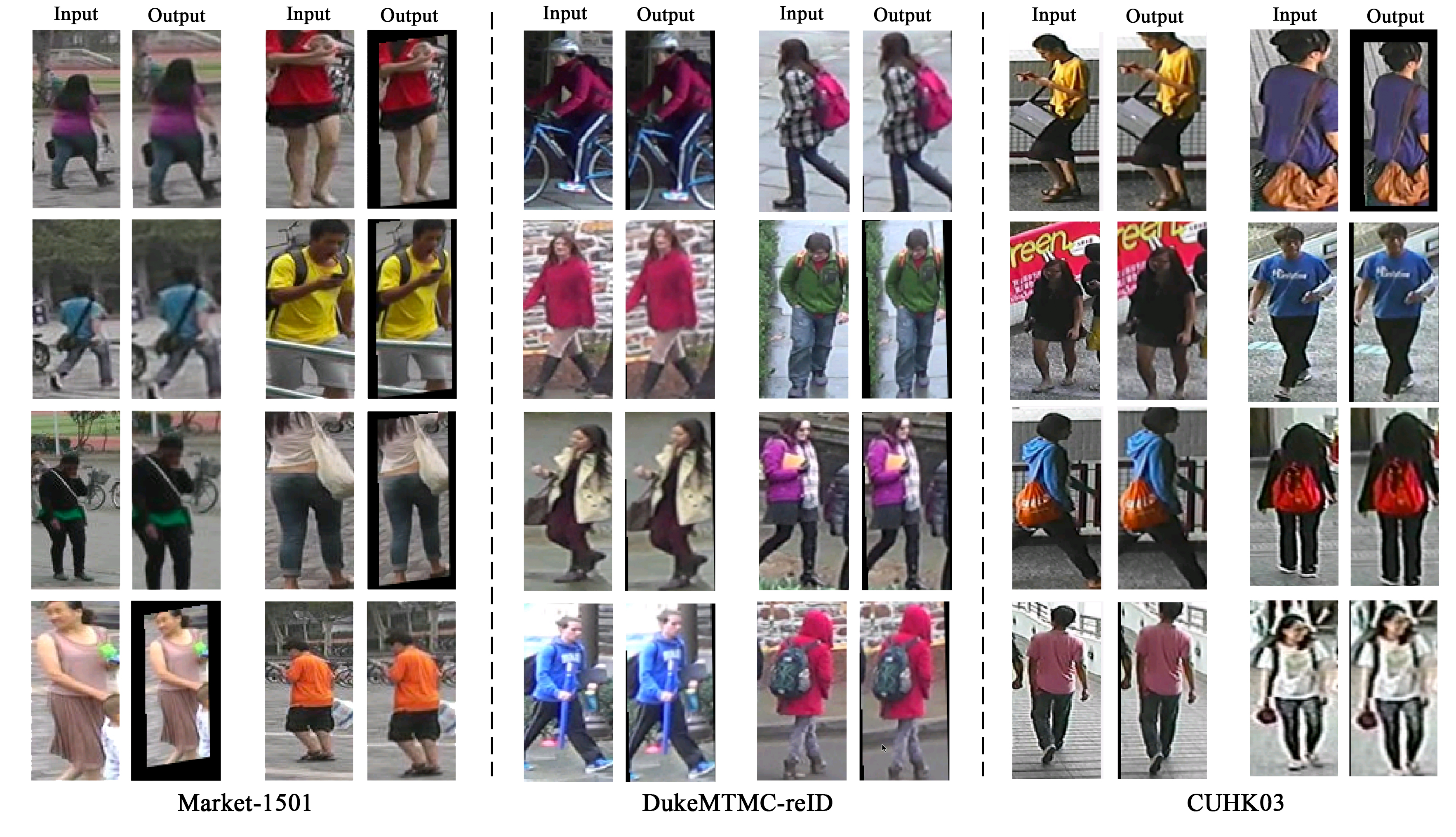}
\end{center}
   \caption{Examples of pedestrian images before and after alignment on three datasets (Market-1501, DukeMTMC-reID and CUHK03). Pairs of input images and aligned images are shown. By removing excessive background or padding zeros to image borders, we observe that PAN reduces the scale and location variance.}
\label{fig:align_sample}
\end{figure*}

\textbf{Visualization of the alignment.} We further visualize the aligned images in Fig. \ref{fig:align_sample}. As aforementioned, the proposed network does not process the alignment on the original image. To visualize the aligned images, we extract the predicted affine parameters and then apply the affine transformation on the originally detected images manually. \textbf{We observe that the network does not perform perfect alignment as the human, but it more or less reduces the scale and position variance, which is critical for the network to learn the representations.} So the proposed network improves the performance of the person re-ID.


\section{Conclusion} \label{conclusion}
Pedestrian alignment and re-identification are two inner-connected problems, which inspires us to develop an attention-based system. In this work, we propose the pedestrian alignment network (PAN), which simultaneously aligns the pedestrians within bounding boxes and learns the pedestrian descriptors. Taking advantage of the attention of CNN feature maps to the human body, PAN addresses the misalignment problem and person re-ID together and thus, improves the person re-ID accuracy. Except for the identity label, we do not need any extra annotation. We also observe that the manually cropped images are not as perfect as preassumed to be. Our network also improves the re-ID performance on the datasets with hand-drawn bounding boxes. Experiments on three different datasets indicate that our method is competitive with the state-of-the-art methods. In the future, we will continue to investigate the attention-based model and apply our model to other fields \ie car recognition. 


\bibliographystyle{apalike} 
\bibliography{mybib}   

\begin{thebibliography}{}

\bibitem[Ahmed et~al., 2015]{ahmed2015improved}
Ahmed, E., Jones, M., and Marks, T.~K. (2015).
\newblock An improved deep learning architecture for person re-identification.
\newblock In {\em CVPR}.

\bibitem[Baltieri et~al., 2015]{baltieri2015mapping}
Baltieri, D., Vezzani, R., and Cucchiara, R. (2015).
\newblock Mapping appearance descriptors on 3d body models for people
  re-identification.
\newblock {\em IJCV}.

\bibitem[Barbosa et~al., 2017]{barbosa2017looking}
Barbosa, I.~B., Cristani, M., Caputo, B., Rognhaugen, A., and Theoharis, T.
  (2017).
\newblock Looking beyond appearances: Synthetic training data for deep cnns in
  re-identification.
\newblock {\em arXiv:1701.03153}.

\bibitem[Chen et~al., 2016]{chen2016similarity}
Chen, D., Yuan, Z., Chen, B., and Zheng, N. (2016).
\newblock Similarity learning with spatial constraints for person
  re-identification.
\newblock In {\em CVPR}.

\bibitem[Chen et~al., 2017]{chen2017exemplar}
Chen, D., Yuan, Z., Wang, J., Chen, B., Hua, G., and Zheng, N. (2017).
\newblock Exemplar-guided similarity learning on polynomial kernel feature map
  for person re-identification.
\newblock {\em IJCV}.

\bibitem[Cheng et~al., 2016]{cheng2016person}
Cheng, D., Gong, Y., Zhou, S., Wang, J., and Zheng, N. (2016).
\newblock Person re-identification by multi-channel parts-based cnn with
  improved triplet loss function.
\newblock In {\em CVPR}.

\bibitem[Cheng et~al., 2011]{cheng2011custom}
Cheng, D.~S., Cristani, M., Stoppa, M., Bazzani, L., and Murino, V. (2011).
\newblock Custom pictorial structures for re-identification.
\newblock In {\em BMVC}.

\bibitem[Deng et~al., 2009]{deng2009imagenet}
Deng, J., Dong, W., Socher, R., Li, L.-J., Li, K., and Fei-Fei, L. (2009).
\newblock Imagenet: A large-scale hierarchical image database.
\newblock In {\em CVPR}.

\bibitem[Ding et~al., 2015]{ding2015deep}
Ding, S., Lin, L., Wang, G., and Chao, H. (2015).
\newblock Deep feature learning with relative distance comparison for person
  re-identification.
\newblock {\em Pattern Recognition}, 48(10):2993--3003.

\bibitem[Felzenszwalb et~al., 2008]{felzenszwalb2008discriminatively}
Felzenszwalb, P., McAllester, D., and Ramanan, D. (2008).
\newblock A discriminatively trained, multiscale, deformable part model.
\newblock In {\em CVPR}.

\bibitem[Geng et~al., 2016]{geng2016deep}
Geng, M., Wang, Y., Xiang, T., and Tian, Y. (2016).
\newblock Deep transfer learning for person re-identification.
\newblock {\em arXiv:1603.06765}.

\bibitem[Girshick, 2015]{girshick2015fast}
Girshick, R. (2015).
\newblock Fast r-cnn.
\newblock In {\em ICCV}.

\bibitem[Gray et~al., 2007]{gray2007evaluating}
Gray, D., Brennan, S., and Tao, H. (2007).
\newblock Evaluating appearance models for recognition, reacquisition, and
  tracking.
\newblock In {\em PETS}, volume~3. Citeseer.

\bibitem[Gray and Tao, 2008]{gray2008viewpoint}
Gray, D. and Tao, H. (2008).
\newblock Viewpoint invariant pedestrian recognition with an ensemble of
  localized features.
\newblock In {\em ECCV}.

\bibitem[He et~al., 2016]{he2016deep}
He, K., Zhang, X., Ren, S., and Sun, J. (2016).
\newblock Deep residual learning for image recognition.
\newblock In {\em CVPR}.

\bibitem[Hermans et~al., 2017]{hermans2017defense}
Hermans, A., Beyer, L., and Leibe, B. (2017).
\newblock In defense of the triplet loss for person re-identification.
\newblock {\em arXiv:1703.07737}.

\bibitem[Hochreiter and Schmidhuber, 1997]{hochreiter1997long}
Hochreiter, S. and Schmidhuber, J. (1997).
\newblock Long short-term memory.
\newblock {\em Neural computation}, 9(8):1735--1780.

\bibitem[Huang et~al., 2012]{huang2012learning}
Huang, G., Mattar, M., Lee, H., and Learned-Miller, E.~G. (2012).
\newblock Learning to align from scratch.
\newblock In {\em NIPS}.

\bibitem[Huang et~al., 2007]{huang2007unsupervised}
Huang, G.~B., Jain, V., and Learned-Miller, E. (2007).
\newblock Unsupervised joint alignment of complex images.
\newblock In {\em ICCV}.

\bibitem[Jaderberg et~al., 2015]{jaderberg2015spatial}
Jaderberg, M., Simonyan, K., Zisserman, A., et~al. (2015).
\newblock Spatial transformer networks.
\newblock In {\em NIPS}.

\bibitem[Johnson et~al., 2015]{johnson2015densecap}
Johnson, J., Karpathy, A., and Fei-Fei, L. (2015).
\newblock Densecap: Fully convolutional localization networks for dense
  captioning.
\newblock {\em arXiv:1511.07571}.

\bibitem[K{\"o}stinger et~al., 2012]{kostinger2012large}
K{\"o}stinger, M., Hirzer, M., Wohlhart, P., Roth, P.~M., and Bischof, H.
  (2012).
\newblock Large scale metric learning from equivalence constraints.
\newblock In {\em CVPR}.

\bibitem[Krizhevsky et~al., 2012]{krizhevsky2012imagenet}
Krizhevsky, A., Sutskever, I., and Hinton, G.~E. (2012).
\newblock Imagenet classification with deep convolutional neural networks.
\newblock In {\em NIPS}.

\bibitem[Li and Wang, 2013]{li2013locally}
Li, W. and Wang, X. (2013).
\newblock Locally aligned feature transforms across views.
\newblock In {\em CVPR}.

\bibitem[Li et~al., 2012]{li2012human}
Li, W., Zhao, R., and Wang, X. (2012).
\newblock Human reidentification with transferred metric learning.
\newblock In {\em ACCV}.

\bibitem[Li et~al., 2014]{li2014deepreid}
Li, W., Zhao, R., Xiao, T., and Wang, X. (2014).
\newblock Deepreid: Deep filter pairing neural network for person
  re-identification.
\newblock In {\em CVPR}.

\bibitem[Liao et~al., 2015]{liao2015person}
Liao, S., Hu, Y., Zhu, X., and Li, S.~Z. (2015).
\newblock Person re-identification by local maximal occurrence representation
  and metric learning.
\newblock In {\em CVPR}.

\bibitem[Lin et~al., 2017]{lin2017improving}
Lin, Y., Zheng, L., Zheng, Z., Wu, Y., and Yang, Y. (2017).
\newblock Improving person re-identification by attribute and identity
  learning.
\newblock {\em arXiv:1703.07220}.

\bibitem[Liu et~al., 2016a]{liu2016end}
Liu, H., Feng, J., Qi, M., Jiang, J., and Yan, S. (2016a).
\newblock End-to-end comparative attention networks for person
  re-identification.
\newblock {\em arXiv:1606.04404}.

\bibitem[Liu et~al., 2016b]{liu2016multi}
Liu, J., Zha, Z.-J., Tian, Q., Liu, D., Yao, T., Ling, Q., and Mei, T. (2016b).
\newblock Multi-scale triplet cnn for person re-identification.
\newblock In {\em ACM Multimedia}.

\bibitem[Liu et~al., 2016c]{liu2016fully}
Liu, X., Xia, T., Wang, J., Yang, Y., Zhou, F., and Lin, Y. (2016c).
\newblock Fully convolutional attention networks for fine-grained recognition.
\newblock {\em arXiv:1611.05244}.

\bibitem[Loy et~al., 2010]{loy2010time}
Loy, C.~C., Xiang, T., and Gong, S. (2010).
\newblock Time-delayed correlation analysis for multi-camera activity
  understanding.
\newblock {\em IJCV}.

\bibitem[Mignon and Jurie, 2012]{mignon2012pcca}
Mignon, A. and Jurie, F. (2012).
\newblock Pcca: A new approach for distance learning from sparse pairwise
  constraints.
\newblock In {\em CVPR}.

\bibitem[Prosser et~al., 2010]{prosser2010person}
Prosser, B., Zheng, W.-S., Gong, S., Xiang, T., and Mary, Q. (2010).
\newblock Person re-identification by support vector ranking.
\newblock In {\em BMVC}.

\bibitem[Qin et~al., 2011]{qin2011hello}
Qin, D., Gammeter, S., Bossard, L., Quack, T., and Van~Gool, L. (2011).
\newblock Hello neighbor: Accurate object retrieval with k-reciprocal nearest
  neighbors.
\newblock In {\em CVPR}.

\bibitem[Ristani et~al., 2016]{ristani2016MTMC}
Ristani, E., Solera, F., Zou, R., Cucchiara, R., and Tomasi, C. (2016).
\newblock Performance measures and a data set for multi-target, multi-camera
  tracking.
\newblock In {\em ECCVW}.

\bibitem[Su et~al., 2016]{su2016deep}
Su, C., Zhang, S., Xing, J., Gao, W., and Tian, Q. (2016).
\newblock Deep attributes driven multi-camera person re-identification.
\newblock {\em ECCV}.

\bibitem[Sun et~al., 2017]{sun2017svdnet}
Sun, Y., Zheng, L., Deng, W., and Wang, S. (2017).
\newblock Svdnet for pedestrian retrieval.
\newblock {\em arXiv:1703.05693}.

\bibitem[Ustinova et~al., 2015]{ustinova2015multiregion}
Ustinova, E., Ganin, Y., and Lempitsky, V. (2015).
\newblock Multiregion bilinear convolutional neural networks for person
  re-identification.
\newblock {\em arXiv:1512.05300}.

\bibitem[Varior et~al., 2016a]{varior2016gated}
Varior, R.~R., Haloi, M., and Wang, G. (2016a).
\newblock Gated siamese convolutional neural network architecture for human
  re-identification.
\newblock In {\em ECCV}.

\bibitem[Varior et~al., 2016b]{varior2016siamese}
Varior, R.~R., Shuai, B., Lu, J., Xu, D., and Wang, G. (2016b).
\newblock A siamese long short-term memory architecture for human
  re-identification.
\newblock In {\em ECCV}.

\bibitem[Vedaldi and Lenc, 2015]{vedaldi15matconvnet}
Vedaldi, A. and Lenc, K. (2015).
\newblock Matconvnet -- convolutional neural networks for matlab.
\newblock In {\em ACM Multimedia}.

\bibitem[Wang and Li, 2012]{wang2012query}
Wang, J. and Li, S. (2012).
\newblock Query-driven iterated neighborhood graph search for large scale
  indexing.
\newblock In {\em ACM Multimedia}.

\bibitem[Wu et~al., 2016a]{wu2016deep}
Wu, L., Shen, C., and Hengel, A. v.~d. (2016a).
\newblock Deep linear discriminant analysis on fisher networks: A hybrid
  architecture for person re-identification.
\newblock {\em arXiv:1606.01595}.

\bibitem[Wu et~al., 2016b]{wu2016personnet}
Wu, L., Shen, C., and Hengel, A. v.~d. (2016b).
\newblock Personnet: Person re-identification with deep convolutional neural
  networks.
\newblock {\em arXiv:1601.07255}.

\bibitem[Wu et~al., 2016c]{wu2016enhanced}
Wu, S., Chen, Y.-C., Li, X., Wu, A.-C., You, J.-J., and Zheng, W.-S. (2016c).
\newblock An enhanced deep feature representation for person re-identification.
\newblock In {\em WACV}.

\bibitem[Xiao et~al., 2016]{xiao2016learning}
Xiao, T., Li, H., Ouyang, W., and Wang, X. (2016).
\newblock Learning deep feature representations with domain guided dropout for
  person re-identification.
\newblock In {\em CVPR}.

\bibitem[Xiao et~al., 2017]{xiao2017joint}
Xiao, T., Li, S., Wang, B., Lin, L., and Wang, X. (2017).
\newblock Joint detection and identification feature learning for person
  search.
\newblock {\em arXiv preprint arXiv:1604.01850}.

\bibitem[Ye et~al., 2015]{ye2015ranking}
Ye, M., Liang, C., Wang, Z., Leng, Q., and Chen, J. (2015).
\newblock Ranking optimization for person re-identification via similarity and
  dissimilarity.
\newblock In {\em ACM Multimedia}.

\bibitem[Yi et~al., 2014]{yi2014deep}
Yi, D., Lei, Z., Liao, S., and Li, S.~Z. (2014).
\newblock Deep metric learning for person re-identification.
\newblock In {\em ICPR}.

\bibitem[Zhang et~al., 2016]{zhang2016learning}
Zhang, L., Xiang, T., and Gong, S. (2016).
\newblock Learning a discriminative null space for person re-identification.
\newblock {\em CVPR}.

\bibitem[Zhao et~al., 2013a]{zhao2013person}
Zhao, R., Ouyang, W., and Wang, X. (2013a).
\newblock Person re-identification by salience matching.
\newblock In {\em ICCV}.

\bibitem[Zhao et~al., 2013b]{zhao2013unsupervised}
Zhao, R., Ouyang, W., and Wang, X. (2013b).
\newblock Unsupervised salience learning for person re-identification.
\newblock In {\em CVPR}.

\bibitem[Zhao et~al., 2014]{zhao2014learning}
Zhao, R., Ouyang, W., and Wang, X. (2014).
\newblock Learning mid-level filters for person re-identification.
\newblock In {\em CVPR}.

\bibitem[Zheng et~al., 2016a]{zheng2016mars}
Zheng, L., Bie, Z., Sun, Y., Wang, J., Su, C., Wang, S., and Tian, Q. (2016a).
\newblock Mars: A video benchmark for large-scale person re-identification.
\newblock In {\em ECCV}.

\bibitem[Zheng et~al., 2017a]{zheng2017pose}
Zheng, L., Huang, Y., Lu, H., and Yang, Y. (2017a).
\newblock Pose invariant embedding for deep person re-identification.
\newblock {\em arXiv:1701.07732}.

\bibitem[Zheng et~al., 2015]{zheng2015scalable}
Zheng, L., Shen, L., Tian, L., Wang, S., Wang, J., and Tian, Q. (2015).
\newblock Scalable person re-identification: A benchmark.
\newblock In {\em ICCV}.

\bibitem[Zheng et~al., 2016b]{zheng2016survey}
Zheng, L., Yang, Y., and Hauptmann, A.~G. (2016b).
\newblock Person re-identification: Past, present and future.
\newblock {\em arXiv:1610.02984}.

\bibitem[Zheng et~al., 2016c]{zheng2016person}
Zheng, L., Zhang, H., Sun, S., Chandraker, M., and Tian, Q. (2016c).
\newblock Person re-identification in the wild.
\newblock {\em arXiv:1604.02531}.

\bibitem[Zheng et~al., 2016d]{zheng2016discriminatively}
Zheng, Z., Zheng, L., and Yang, Y. (2016d).
\newblock A discriminatively learned cnn embedding for person
  re-identification.
\newblock {\em arXiv:1611.05666}.

\bibitem[Zheng et~al., 2017b]{zheng2017unlabeled}
Zheng, Z., Zheng, L., and Yang, Y. (2017b).
\newblock Unlabeled samples generated by gan improve the person
  re-identification baseline in vitro.
\newblock {\em arXiv:1701.07717}.

\bibitem[Zhong et~al., 2017]{zhong2017re}
Zhong, Z., Zheng, L., Cao, D., and Li, S. (2017).
\newblock Re-ranking person re-identification with k-reciprocal encoding.
\newblock In {\em CVPR}.

\end{thebibliography}

\end{document}